\documentclass[10pt,twocolumn,letterpaper]{article}

\usepackage{cvpr}
\usepackage{times}
\usepackage{epsfig}
\usepackage{amsmath}
\usepackage{amssymb}
\usepackage{subfigure}
\usepackage{booktabs}
\usepackage{float}
\usepackage{cite}
\usepackage[bottom]{footmisc}

\DeclareFontFamily{U}{dutchcal}{\skewchar\font=45 }
\DeclareFontShape{U}{dutchcal}{m}{n}{<-> s*[1.0] dutchcal-r}{}
\DeclareFontShape{U}{dutchcal}{b}{n}{<-> s*[1.0] dutchcal-b}{}
\DeclareMathAlphabet{\mathlcal}{U}{dutchcal}{m}{n}
\SetMathAlphabet{\mathlcal}{bold}{U}{dutchcal}{b}{n}

\newcommand{\startcompact}[1]{\par\vspace{-0.75em}\begin{#1}%
\allowdisplaybreaks\ignorespaces}
\newcommand{\stopcompact}[1]{\end{#1}\ignorespaces}

\newenvironment{packed_item}{
\begin{itemize}
   \setlength{\itemsep}{1pt}
   \setlength{\parskip}{0pt}
   \setlength{\parsep}{0pt}
 }{\end{itemize}}

\usepackage[pagebackref=true,breaklinks=true,colorlinks,bookmarks=false]{hyperref}

\cvprfinalcopy 

\def\cvprPaperID{1490} 

\begin{document}
\title{Image2Mesh: A Learning Framework for Single Image 3D Reconstruction}

\author{Jhony K. Pontes$^{\dagger}$, Chen Kong$^\star$, Sridha Sridharan$^\dagger$, Simon Lucey$^\star$, Anders Eriksson$^\dagger$, Clinton Fookes$^\dagger$\\
\begin{tabular}{cc}
Queensland University of Technology$^\dagger$ & Carnegie Mellon University$^\star$
\end{tabular}
}
 
\twocolumn[{%
\renewcommand\twocolumn[1][]{#1}%

\maketitle

\vspace{-1cm}
\begin{figure}[H]
\setlength{\hsize}{\textwidth}
\centering
\includegraphics[width=1.01\textwidth,keepaspectratio]{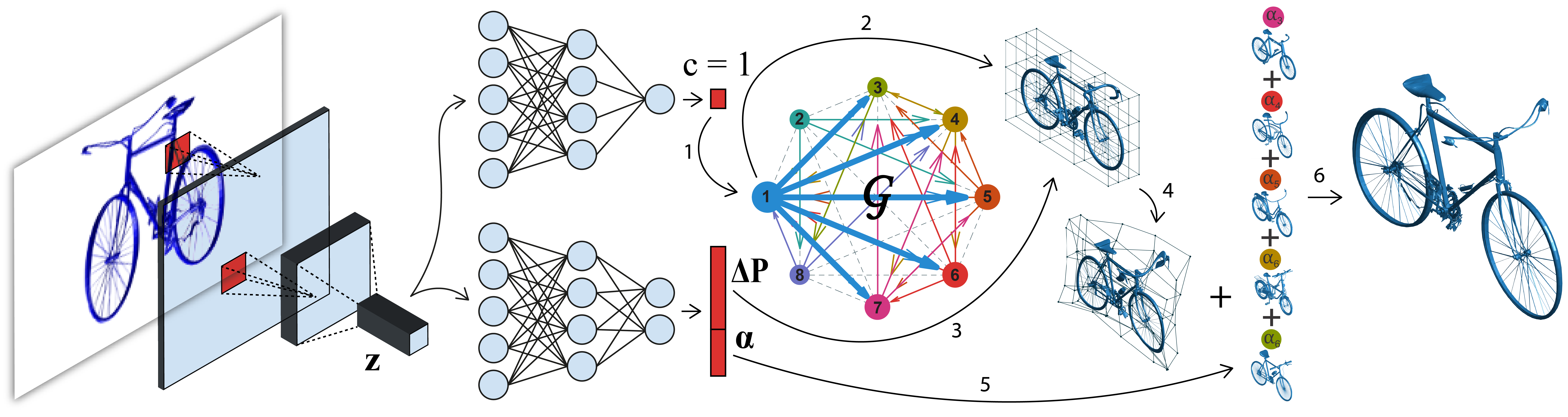} 
\caption{Given a single image, our framework employs a convolutional autoencoder to extract the image's latent space, $\mathbf{z}$, that is used to classify it to an index, $c$, using a multi-label classifier and regress it to a compact shape parametrization using a feedforward network. We use a graph embedding, $\mathcal{G}$, that compactly represents 3D mesh objects to reconstruct the 3D model. Firstly, the estimated index, $c$, selects the closest 3D model to the image from the graph. Secondly, the selected model is deformed through the estimated parameters - free-form deformation (\textsc{Ffd}), $\mathbf{\Delta P}$, and sparse linear combination parameters (\ie $\alpha$'s). In this example, model 1 is selected (arrows 1 and 2), \textsc{Ffd} is then applied (arrows 3 and 4), and finally the linear combination with the nodes 3, 4, 5, 6, and 7 (blue arrows on the graph that indicates the models in dense correspondence with node 1) are performed (arrow 5) to reconstruct the final 3D mesh model (arrow 6).}
\label{fig:overview}
\end{figure}%
}] 


\begin{abstract}
One challenge that remains open in 3D deep learning is how to efficiently represent 3D data to feed deep networks. Recent works have relied on volumetric or point cloud representations, but such approaches suffer from a number of issues such as computational complexity, unordered data, and lack of finer geometry. This paper demonstrates that a mesh representation (\ie vertices and faces to form polygonal surfaces) is able to capture fine-grained geometry for 3D reconstruction tasks. A mesh however is also unstructured data similar to point clouds. We address this problem by proposing a learning framework to infer the parameters of a compact mesh representation rather than learning from the mesh itself. This compact representation encodes a mesh using free-form deformation and a sparse linear combination of models allowing us to reconstruct 3D meshes from single images. In contrast to prior work, we do not rely on silhouettes and landmarks to perform 3D reconstruction. We evaluate our method on synthetic and real-world datasets with very promising results. Our framework efficiently reconstructs 3D objects in a low-dimensional way while preserving its important geometrical aspects.
\end{abstract}

\section{Introduction}
Most of us take for granted the ability to effortlessly perceive our surroundings world and its objects in three dimensions. In general, we have great ideas about the 3D space only by looking at a single 2D image of an object even when there are many possible shapes that could have produced the same image. We simply rely on assumptions and prior knowledge acquired throughout our lives for the inference. It is one of the fundamental goals of computer vision to give machines the ability to perceive its surroundings as we do, for the purpose of providing solutions to tasks such as
self-driving cars, virtual and augmented reality, robotic surgery, to name a few.

A specific problem that is of particular interest towards achieving this ambition of 
human-like machine perception is that of recovering 3D information from a single image. This exceedingly difficult 
and highly ambiguous problem is typically addressed by incorporating prior knowledge about the scene such as shape or 
scene priors \cite{Wang2014,Zhou2015,Zhou2016,Wu2016_2,Chen2016,Bansal2016,Han2016,Chen2017}. 
This body of work has provided a very good foundation for this task and it has in particular indicated that the use of
shape priors is highly beneficial. With the availability of millions of 3D CAD models across different categories 
publicly available online, this becomes even more attractive and motivating. This is a realisation we propose to exploit 
in this work.

With the recent arrival of deep learning many interesting work have been done to tackle 3D inference from a 2D image by exploring the abundance of 3D models available online \cite{choy2016,Sharma2016,Maxim2017,Riegler2017}. Most of them rely on volumetric shape representation, an approach arguably motivated by the ease of which convolutions can be 
generalized from 2D to 3D. A significant drawback these methods have is that the computational and memory costs scale cubically with the resolution. Octrees \cite{Maxim2017} and point cloud \cite{Qi2017} representations have been proposed to make the learning more efficient. However, despite of its improved performance, such representations still fail to capture fine-grained geometry as a dense 3D mesh representation would capture.

The aim of this work is to propose a compact representation of 3D mesh models to enable us to learn its parameters for the single image 3D reconstruction task. We want to exploit a dense mesh representation to better unlock fine-grained geometry. To do this in a scalable and computationally efficient manner, we propose a framework based on a graph that embeds 3D mesh objects in a low-dimensional space and still allow us to achieve compelling reconstructions with exceptionally detailed geometry. We draw inspiration by the works in \cite{Chen2017,Pontes2017}, where a graph embedding to compactly model the intrinsic variation across classes of 3D models was proposed. Any 3D mesh model can be parametrized in terms of free-form deformation (\textsc{Ffd}) \cite{Sederberg1986} and sparse linear representation in a dictionary. \textsc{Ffd} embeds a 3D mesh model in a grid space where deformations can be performed by repositioning a smaller number of control points. ``Free-form'' means that we are able to deform whatever the object is and whatever its topology. 
More importantly, it also conserves the objects visual aspects \cite{Sederberg1986}. 

Our method first classifies the latent space of an image to retrieve a coarse 3D model from the graph. Then, the compact shape parameters are estimated from a feedforward neural network which maps the image features to the shape parameters space - \textsc{Ffd} and sparse linear combination parameters. The dense 3D mesh model is then recovered by applying the estimated deformations to the model selected. An overview of the proposed framework is illustrated in Figure~\ref{fig:overview}.

\noindent\textbf{Our main contributions are:}
\begin{packed_item}
\item We propose a novel learning framework to estimate a dense 3D mesh model through a low-dimensional embedding space from a single image;
\item We quantitatively and qualitatively demonstrate through synthetic data that our method is able to estimate the correct 3D models from a single image.  
\item We show impressive results for dense and realistic 3D reconstructions from real-world images.
\end{packed_item}

\section{Related Work}
Recent advances in neural networks and the availability of readily available 3D models (such as ShapeNet \cite{ShapeNet})
has sparked a considerable interest in methods using deep learning to solve tasks related to geometry and 3D 
reconstruction.
Several papers have proposed a 3D volumetric representation \cite{Wu2015,Ulusoy2015,Cherabier2016,choy2016,Rezende2016,Yan2016,Sharma2016,QiSu2016,Kar2017,Zhu2017,marrnet2017} so they can feed deep neural networks applying 3D convolutions, pooling, and other techniques that have been successfully applied to 2D images for the learning process. Volumetric autoencoders \cite{Girdhar2016,Sharma2016} and generative adversarial networks (GAN) have been proposed \cite{Wu2016,Liu2017,Gwak2017} to learn of the probabilistic latent space of object shapes. It has been used for object completion, classification and reconstruction. Despite of all the great work, volumetric representation has a great drawback. The memory and the computational costs grow cubically as the voxel resolution increases which limits such works to low-resolution 3D reconstructions. 

Octree-based convolutional neural networks have been presented to manage these limitations \cite{Riegler2017,Wang2017,Hane2017,Maxim2017}. It splits the 3D voxel grid by recursively subdividing it into octants thus reducing the computational complexity of the 3D convolution. The computations are then focused on regions where most of the information about the object's geometry is contained - normally on its surface. Although it allows for higher resolution outputs, around $64^3$ voxels, and a more efficient training, the 3D volumetric models still lacks fine-scaled geometry. In trying to provide some answers to these shortcomings, a more efficient input representation for 3D geometry using point clouds have been recently entertained \cite{Su2017,Qi_CVPR2017,Qi2017,lin2017learning}. PointNet \cite{Qi2017} is able to learn straight from unstructured points as inputs by employing max pooling to generate feature descriptors of the original input. Point cloud generators might be a good solution but such architectures have been shown to generate relatively low-resolution reconstruction. It does not encode detailed geometrical information yet. In fact, obtaining a realistic 3D model with compelling visual aesthetics from unstructured point clouds is a difficult problem.

Reconstruction of surfaces from unstructured point clouds is a highly challenging problem, with applications mainly in computer graphics, and it can be even more problematic in case of incomplete, noisy and sparse data \cite{Nan2017}. On the other hand, 3D shapes can be efficiently represented by polygon meshes. However, it is difficult to parametrize meshes to be used with convolutional neural networks since it has also an irregular pattern \cite{Lun3DV2017}. A deep residual network to generate 3D surfaces has been proposed in \cite{Sinha2017}. The limitation however is that it can only manage simple surfaces (\ie genus-0 surface). A novel graph embedding based on the local dense correspondences between 3D mesh models has been proposed in \cite{Chen2017,Pontes2017}. The method is able to reconstruct the finer geometry of a single image based on a low-dimensional parametrization based on \textsc{Ffd} and sparse linear combination. Although they showed impressive results for dense and realistic 3D reconstruction through mesh models, it relies on the image silhouette and a few semantic landmarks to perform the geometrical optimization. Our method employs the same compact mesh representation but without relying on silhouettes and landmarks for the dense 3D reconstruction.

\section{Proposed Learning Framework}
Our aim is to learn an algorithm that, given a single image from a specific category (\eg bicycle, car, chair, etc.), has the ability to estimate a dense 3D mesh model. Our framework takes advantage of the compact 3D mesh representation proposed in \cite{Chen2017,Pontes2017} but it estimates this representation directly from the image and without the use of silhouettes or known landmarks which heavily constrain its applicability. It encodes a mesh model in a low-dimensional parametrization composed of a model index, \textsc{Ffd} displacements (\eg 32 control points), and sparse linear combination weights which depends on the size of the graph/dictionary of CAD models. 
We train a classifier to estimate the index and a feedforward neural network to regress the \textsc{Ffd} and the sparse linear combination parameters from the latent space of an image learnt from a convolutional autoencoder (CAE). A model is then selected from the graph using the estimated index and the \textsc{Ffd} parameters are applied to initially deform the model. Once a satisfactory model candidate is selected and deformed, we have, from the graph, the information about what models are possible to establish dense correspondences. Thus, we can apply the sparse linear combination parameters to refine the object and get the final 3D mesh model. The framework is shown in Figure~\ref{fig:overview}. 

\subsection{3D Mesh Parametrization}
\label{3dparams}
We parametrize a 3D mesh model using the compact shape representation method presented in \cite{Chen2017,Pontes2017}. It uses a graph $\mathcal{G}$ with 3D mesh models from the same class as nodes and its vertices indicates if we can establish dense correspondences or not (see an example in Figure~\ref{fig:overview}). Note that the graph is not fully connected but sparse. So in every subgraph we can perform sparse linear combinations to deform a 3D model (\ie a union of subspaces). Mathematically, consider $\Omega$ as an index set of the nodes in a certain subgraph with $\mathcal{S}_c(\mathbf{V}_c,\mathbf{F}_c)$ as the central shape node. $\mathbf{V}$ and $\mathbf{F}$ stand for the shape vertices and faces. Dense correspondences will always exist for all $i \in \Omega$ and this allows us to deform the model $\mathcal{S}(\mathbf{V},\mathbf{F})$ by linear combination,

\startcompact{small}
\begin{equation}
	\begin{aligned}
		\mathbf{V} = \alpha_c \mathbf{V}_c + \sum_{i \in \Omega} \alpha_i \mathbf{V}_c^i, \quad \mathbf{F} = \mathbf{F}_c,
	\end{aligned}
\label{eq:linearComb}
\end{equation}
\stopcompact{small}

\noindent where $\alpha$'s are weights for the linear combination. A two-step process is then performed, first we need to find a candidate model (\ie a node) from $\mathcal{G}$. Second, knowing the index that indicates the central node we can deform the model by linear interpolating it with its dense correspondences from the subgraph. To select a good candidate in the first step, \textsc{Ffd} is used to deform a model and pick the one which best fit an image. A mesh is represented in terms of \textsc{Ffd} as,

\startcompact{small}
\begin{equation}
	\mathbf{S}_\mathit{ffd} = \mathbf{B \Phi (P + \Delta P)},
 	\label{eq:FFD}
\end{equation}
\stopcompact{small}

\noindent where $\mathbf{S}_\mathit{ffd} \in \mathbb{R}^{N \times 3}$ are the vertices of the 3D mesh, $\mathbf{B} \in \mathbb{R}^{N \times M}$ is the deformation matrix, $\mathbf{P} \in \mathbb{R}^{M \times 3}$ are the control point coordinates, $N$ and $M$ are the number of vertices and control points respectively, $\mathbf{\Delta P}$ are the control point displacements, and $\mathbf{\Phi}$ is a matrix to impose symmetry in the \textsc{Ffd} grid as in \cite{Pontes2017}. The deformation matrix $\mathbf{B}$ is a set of polynomial basis (in our case Bernstein polynomials). See \cite{Chen2017,Pontes2017} for detailed information about the graph creation and deformation process.

Having the graph, we can embed a 3D mesh model in a low-dimensional space. We only need an index that indicates what model we should pick up from the graph, the symmetric \textsc{Ffd} displacements, $\mathbf{\Delta P}$, to apply the initial deformation in the selected model, and the sparse linear combination weights, $\mathbf{\alpha}$. The linear interpolation between objects in a subgraph refines the final 3D model. In this work, we asked ourselves whether we could learn those low-dimensional parameters by regressing them from a single image to reconstruct 3D mesh models.

\begin{figure*}[ht!]
\includegraphics[trim={0cm 0.5cm 0cm 1.3cm},clip,width=\textwidth,height=8cm,keepaspectratio]{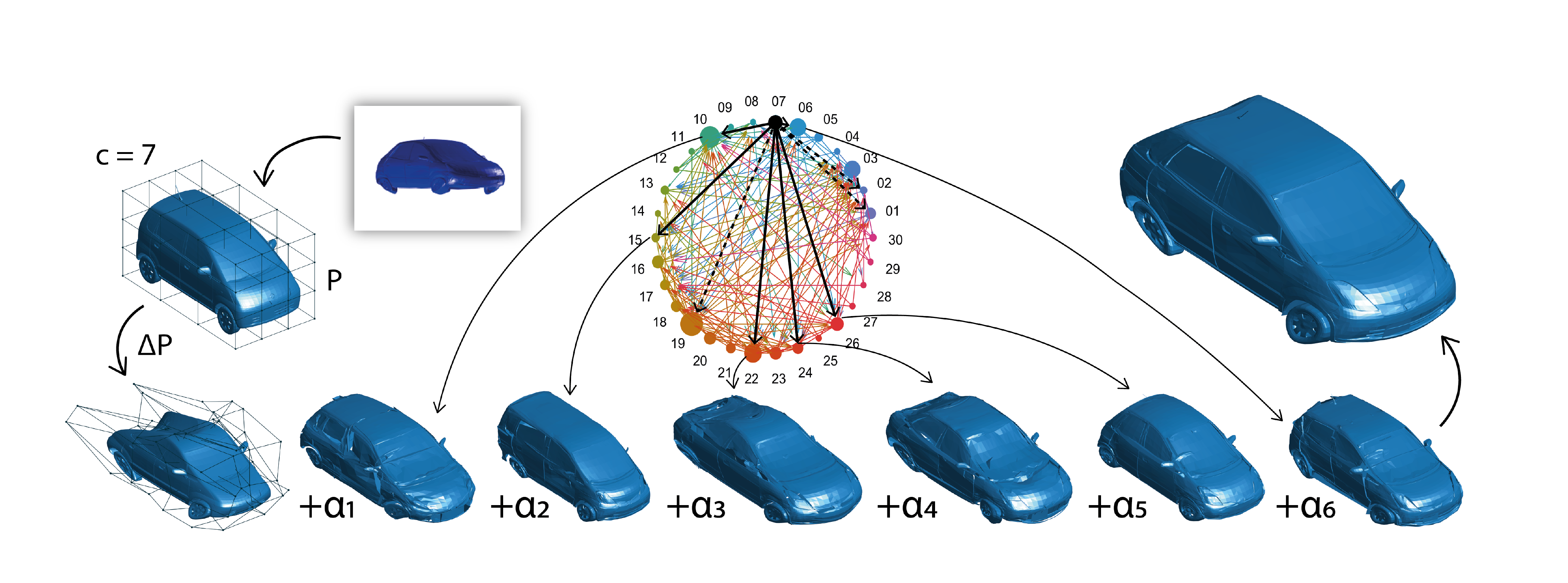}
\caption{Given an input image and a graph of a car, for instance, with 30 CAD models as nodes, our method first selected the model/node 7 from the graph. Then it is deformed by applying the \textsc{Ffd} displacements, $\mathbf{\Delta P}$, on the initial \textsc{Ffd} grid, $\mathbf{P}$. Afterwards, the linear combination is performed with the estimated $\mathbf{\alpha}$ to reconstruct the final model. The black arrows on the graph representation show what models are possible to perform linear combination with the model selected (\ie the models with dense correspondences). Note that not all the models were selected, but 6 out of 9 models in this example. (The size of the nodes in the graph varies to illustrate the number of edges starting from it. A big node means that we have more dense correspondences than a smaller node.)}
\label{fig:fig2}
\end{figure*}

\subsection{Collecting Synthetic Data}
\label{data}
Since we aim to learn 3D mesh models from a single image using the compact shape representation proposed in \cite{Pontes2017}, we need to synthetically generate data containing 3D models, its compact parametrization (\textsc{Ffd} + $\mathbf{\alpha}$ parameters), and rendered images. 
One may question why we do not simply parametrize the whole ShapeNet, for example, using the shape embedding proposed in \cite{Pontes2017}. One of the drawbacks of such an approach would be that it relies on 3D semantic landmarks that were manually annotated in some CAD models from ShapeNet to obtain a compact shape embedding. Because of this, we decided to use the graph to generate data for each object class instead of manually annotating 3D anchors on several models which is laborious. 

To generate the data, we randomly choose an index and then we deform the selected model by applying a $\mathbf{\Delta P}$ and $\mathbf{\alpha}$ from a learnt probability density function (PDF). To learn a PDF for the displacements, $\mathbf{\Delta P}$, we use a Gaussian Mixture Model (GMM) to capture information about the nature of the desired deformations. Since we have, from the graph creation process, the \textsc{Ffd} parameters for every pair of 3D models in the graph (got during the deformation process to find dense correspondences in \cite{Pontes2017}), we can learn a GMM using such prior information. For the sparse linear coefficients' PDF we simply fit a normal distribution to a set of $\alpha$'s from 3D reconstructions got from \cite{Pontes2017} on the PASCAL3D+ dataset \cite{Pascal3D}. Armed with this, we can synthetically generate deformed 3D mesh models and images with their respective low-dimensional shape parametrizations.

\subsection{Learning the Image Latent Space}
\label{cae}
To obtain a lower-dimensional representation for the images, a convolutional autoencoder is proposed to extract useful features in an unsupervised manner. The encoder network consists of three fully convolution layers with numbers of kernels $\{8,16,32\}$, kernel sizes $\{5,3,3\}$, and strides $\{3,3,3\}$. The decoder network has three fully transposed convolution layers with numbers of kernels $\{16,8,1\}$, kernel sizes $\{3,3,5\}$, and strides $\{3,3,3\}$. The input takes grayscale images of size $220 \times 220$. All layers use ReLU as activation functions except the last one that uses Tanh. This gives us a network flow of size $220^2 \to 72^2 \to 24^2 \to 8^2 \to 24^2 \to 72^2 \to 220^2$, respectively. We take the image latent space, $\mathbf{z} \in \mathbb{R}^{2,048}$, from the last layer of the encoder as feature representation.

\subsection{Learning the Index to Select a Model}
Firstly we need to retrieve a 3D model from the graph (a graph is explained in Subsection~\ref{3dparams}). We treat that as a multi-label classification problem. For example, if a graph of the class `car' would have 30 different cars/nodes it would have 30 indices/labels (1 to 30) that an image of a car could be classified. In this way, the ``closest" car to the image can be selected from the graph according to the estimated index. For this purpose, we propose a simple yet effective multi-label classifier to estimate graph indices. The input is the image latent space from the convolutional autoencoder of size 2,048 ($8 \times 3 \times 3$). The output is a one-hot encoded vector that represents the ground truth indices. The network has one hidden layer of size 1,050 and ReLU is used as activation function.

\subsection{Learning the Shape Parameters}
We wish to learn a mapping from the image feature representation $\mathbf{z}$ of size 2,048 to its corresponding 3D shape parameters, $\mathbf{\Delta P}$ and $\mathbf{\alpha}$, $f:\mathbf{z} \to \{\mathbf{\Delta P}, \mathbf{\alpha}\}$. Given the training set of image features and the ground truth shape parameters, we learn this mapping using a feedforward neural network. The input is the image latent space of size 2,048. The output is a vector containing the shape parameters, $\mathbf{\kappa}  \in \mathbb{R}^{M N}$, where $M$ and $N$ are the number of \textsc{Ffd} and sparse linear combination parameters, $\mathbf{\alpha}$, respectively. For instance, $\mathbf{\kappa}  \in \mathbb{R}^{126}$, where 96 values would be the \textsc{Ffd} parameters $(32 \times 3)$ and the remaining 30 values would be the $\mathbf{\alpha}$ parameters for the sparse linear combination of models in the graph. The network has one hidden layer of size 1,500 and it uses ReLU as activation function.

\section{Experiments}

\begin{table*}
\centering
\resizebox{0.935\textwidth}{!}{\begin{tabular}{@{}rccccccccccccc@{}} 
\toprule
&\multicolumn{2}{c}{\textbf{CAE}} &\hphantom &\multicolumn{4}{c}{\textbf{3D Model Selection}} &\hphantom &\multicolumn{2}{c}{\textbf{Params Estimation}} &\hphantom &\multicolumn{2}{c}{\textbf{3D Reconstruction}} \\
\cmidrule{2-3} \cmidrule{5-8} \cmidrule{10-11} \cmidrule{13-14}
&$MSE$ &$\sim t$ (min) &&$Acc(\%)$ &$Prec(\%)$ &$Rec(\%)$ &$\sim  t$ (min) &&$MSE$ &$\sim t$ (min) &&$dist_{3D}$ &$IoU$ \\ \midrule

\textbf{car} 		&\textbf{0.0012} &30   &&92.13 &92.11 &93.33 &76    &&0.0176   &197    &&0.006  &0.664 \\
\textbf{bicycle} 	&0.0068 &25   &&92.07 &92.00 &92.18 &73    &&0.0102   &208    &&0.025  &0.795 \\
\textbf{motorbike}	&0.0045 &21   &&\textbf{97.80} &\textbf{97.87} &\textbf{97.72} &74    &&0.0150   &322    &&0.007  &0.679 \\
\textbf{aeroplane}	&0.0017 &24   &&77.33 &76.90 &78.09 &96    &&0.0108   &209	  &&0.023  &0.551 \\
\textbf{bus}		&0.0013 &35   &&90.20 &90.05 &92.14 &74    &&0.0329   &156    &&\textbf{0.003}  &\textbf{0.776} \\
\textbf{chair}		&0.0022 &21   &&87.33 &86.97 &88.74 &75    &&\textbf{0.0090}   &158    &&0.034  &0.403	\\
\textbf{diningtable}&0.0020 &22   &&73.60 &73.60 &74.98 &75    &&0.0119   &158    &&0.165  &0.332	\\
\textbf{sofa}		&0.0013 &21   &&76.20 &76.47 &77.69 &74    &&0.0119   &156    &&0.063  &0.402 \\ \midrule
\textbf{Mean}		&0.0024	&25	  &&85.86 &85.75 &86.86 &77    &&0.0144   &195 	  &&0.041 			 &0.575 \\	

\bottomrule
\end{tabular}}
\caption{Quantitative results of our learning framework evaluated on the synthetic test set. We show the performance of the convolutional autoencoder (CAE), the multi-label classification for the 3D model selection, the feedforward network for the parameters estimation, and the 3D reconstruction from single image. $Acc$, $Prec$, $Rec$, and $t$ stand for the accuracy, precision, recall, and the training time, respectively.}
\label{tab:results}
\end{table*}

\textbf{Datasets.} To train our framework we take the approach of synthesizing 3D mesh models using the strategy discussed in the Subsection~\ref{data}. We generated 5,000 deformed 3D models of eight object categories (car, bicycle, motorbike, aeroplane, bus, chair, diningtable, and sofa) using the graphs from \cite{Pontes2017}. Every graph from \cite{Pontes2017} has 30 CAD models sampled from ShapeNet \cite{ShapeNet} except for the bicycle and motorbike graphs that have 21 and 27 CAD models, respectively. We rendered for every 3D synthesised model a 2D view of size $256 \times 192$ using different viewpoints with a white background. We also produced uniform lighting across the surfaces of the object. With the images and the ground truth 3D meshes, indices and shape parameters, we can train our framework and evaluate the performance. The data was split in $70\%$ for training and $30\%$ for testing.

\textbf{Evaluation metrics.} To quantify the quality of the classification step we employed the accuracy, precision and recall metrics. To evaluate the estimated shape parameters we use the mean squared error (MSE). For the 3D shape reconstruction measure we use the symmetric surface distance, $s_{dist}$, to the ground truth. $s_{dist}$ is computed by densely sampling points on the faces and using normalized points distance to estimate the model similarity, defined as,

\startcompact{small}
\begin{equation}
	\begin{aligned}
		dist_{3D} = \frac{1}{| \mathbf{\hat{V}} |} \sum_{\mathbf{v_i} \in \mathbf{\hat{V}}} dist(\mathbf{v_i}, \mathcal{S}) + \frac{1}{| \mathbf{V} |} \sum_{\mathbf{v_i} \in \mathbf{V}} dist(\mathbf{v_i}, \mathcal{\hat{S}}), 
	\end{aligned}
\label{eq:surf_dist}
\end{equation}
\stopcompact{small}

\noindent where $\mathbf{\hat{V}}$, $\mathcal{\hat{S}}$, $\mathbf{V}$,  $\mathcal{S}$ are the estimated vertices and surfaces, and the ground truth vertices and surfaces, respectively. 

Moreover, we use the intersection over union (IoU) between the voxel models as in \cite{choy2016} defined as,

\startcompact{small}
\begin{equation}
	\begin{aligned}
		s_{IoU} = \frac{\mathcal{\hat{V}} \cap \mathcal{V}}{\mathcal{\hat{V}} \cup \mathcal{V}},
	\end{aligned}
\label{eq:iou}
\end{equation}
\stopcompact{small}

\noindent where $\mathcal{\hat{V}}$ and $\mathcal{V}$ are the voxel models of the estimated and ground truth models respectively.

\subsection{Estimating the Image Latent Space}
The first set of experiments were performed on the CAE to learn a latent representation from an image. We found the architecture described in the Subsection~\ref{cae} to have the better performance. The image feature representation is discriminative and performed well on the classifier and on the feedforward network to estimate the shape parameters. Table~\ref{tab:results} shows the MSE on the test set and the time spent training the network for every class used.

\subsection{Selecting a 3D Mesh Model}
The first stage of our learning framework after having the image latent space is the selection of a mesh model from the given graph. The performance of the proposed multi-label classifier is shown in Table~\ref{tab:results}. One can note that the overall performance on the test set was satisfactory in terms of accuracy (85.68\%), precision (85.75\%) and recall (86.86\%). The best performance was achieved on the motorbike category (27 labels) with an accuracy of 97.80\%. The category have very different motorbikes from each other which explains the great performance. Besides, the diningtable category (30 labels) had the lowest performance, with an accuracy of 73.60\%. This is due to the synthesised images look similar since there are not many unique tables in the object class. Moreover, we fixed the network architecture for all classes. Fine tuning the classifier would for specific classes will most likely improve performance even further. 

\subsection{Estimating the Shape Parameters}
The last bit of our framework before the 3D reconstruction is the estimation of the shape parameters. The results are also shown in Table~\ref{tab:results}. The overall MSE (0.0144) on the testing set shows that the network is indeed learning a mapping function to estimate the \textsc{Ffd} and the $\mathbf{\alpha}$ parameters from the image latent space. The graphs used have about 30 mesh models each which means that once we have selected a model we can establish dense correspondences with up to 29 models (29 $\alpha$ values). The resolution of the \textsc{Ffd} grid is of $4^3$ that gives us 64 control points to free-deform the model. Since the majority of man-made object are symmetric, we impose a symmetry on the \textsc{Ffd} grid so that the deformations are forced to be symmetric and more realistic. Therefore, we have to estimate only half of the \textsc{Ffd} parameters. The feedforward network then maps the image latent space to 32 displacements of the control points in the 3D space, $\mathbf{\Delta P} \in \mathbb{R}^{32 \times 3}$, and to 30 sparse linear combinations parameters, $\mathbf{\alpha} \in \mathbb{R}^{30}$ (29 + the model selected). The estimated parameters, in this case, is of size $\mathbf{\kappa} \in \mathbb{R}^{126}$. One can note that it is a very low-dimensional parametrization that is efficiently learned through a simple network architecture.

\subsection{3D Reconstruction from a Single Image}
Given a single image we can forward pass it to our learned framework to estimate an index, $c$, and the shape parameters, $\mathbf{\kappa}$. A 3D mesh model is initially selected from the class-specific graph by the estimated index, $c$. Afterwards, the \textsc{Ffd} displacements, $\mathbf{\Delta P}$, is applied to free-deform the model using the Equation~\ref{eq:FFD}, $\mathbf{S}_\mathit{ffd} = \mathbf{B \Phi (P + \Delta P)}$. Note that we only need to add the estimated displacements to the initial grid of control points $\mathbf{P}$. Finally, we can apply the linear combination parameters, $\mathbf{\alpha}$, to deform the model through the Equation~\ref{eq:linearComb}, $\mathbf{V} = \alpha_c \mathbf{V}_c + \sum_{i \in \Omega} \alpha_i \mathbf{V}_c^i$.

\textbf{3D Reconstruction from Synthetic Images.} The initial experiments were performed on synthetic images from the 8 classes were we have the ground truth 3D mesh models and also the shape embedding parameters. Figure~\ref{fig:fig2} shows a real example of our framework flow. Given an input image and a graph of a car with 30 CAD models as nodes, our method first selected the model 7 ($c = 7$). Then, the selected model is deformed by applying the \textsc{Ffd} displacements, $\mathbf{\Delta P}$ on the initial \textsc{Ffd} grid, $\mathbf{P}$. Finally, the linear combination is performed with the estimated $\mathbf{\alpha}$ to reconstruct the final model. Note that not all the models were selected to perform the linear combination, but 6 out of 9 models in this example. Table~\ref{tab:results} summarizes the quantitative results of our 3D reconstruction on the synthetic test set. We measured the quality of the 3D reconstruction through the surface distance metric, $dist_{3D}$, and the IoU between the reconstructed and the ground truth model. Our framework clearly performed well on the synthetic dataset according to the surface distance metric. The IoU of the classes aeroplane, chair, diningtable, and sofa had the lowest values which means that a good voxel intersection between the reconstructed model and the ground truth was not possible. This happens, for instance, due to a little shift on the aeroplane wings, chair's legs, etc, so it is not possible to get a good voxel intersection.

\begin{figure*}[ht!]
\centering
\textbf{\hspace{-5pt} (a) Input \hspace{8pt} (b) Selected model \hspace{8pt} (c) \textsc{Ffd} \hspace{13pt} (d) Final model \hspace{10pt} (e) Voxel model \hspace{15pt} (f) GT}
\includegraphics[trim={0cm 1cm 1.8cm 0.5cm},clip,width=0.85\textwidth,keepaspectratio]{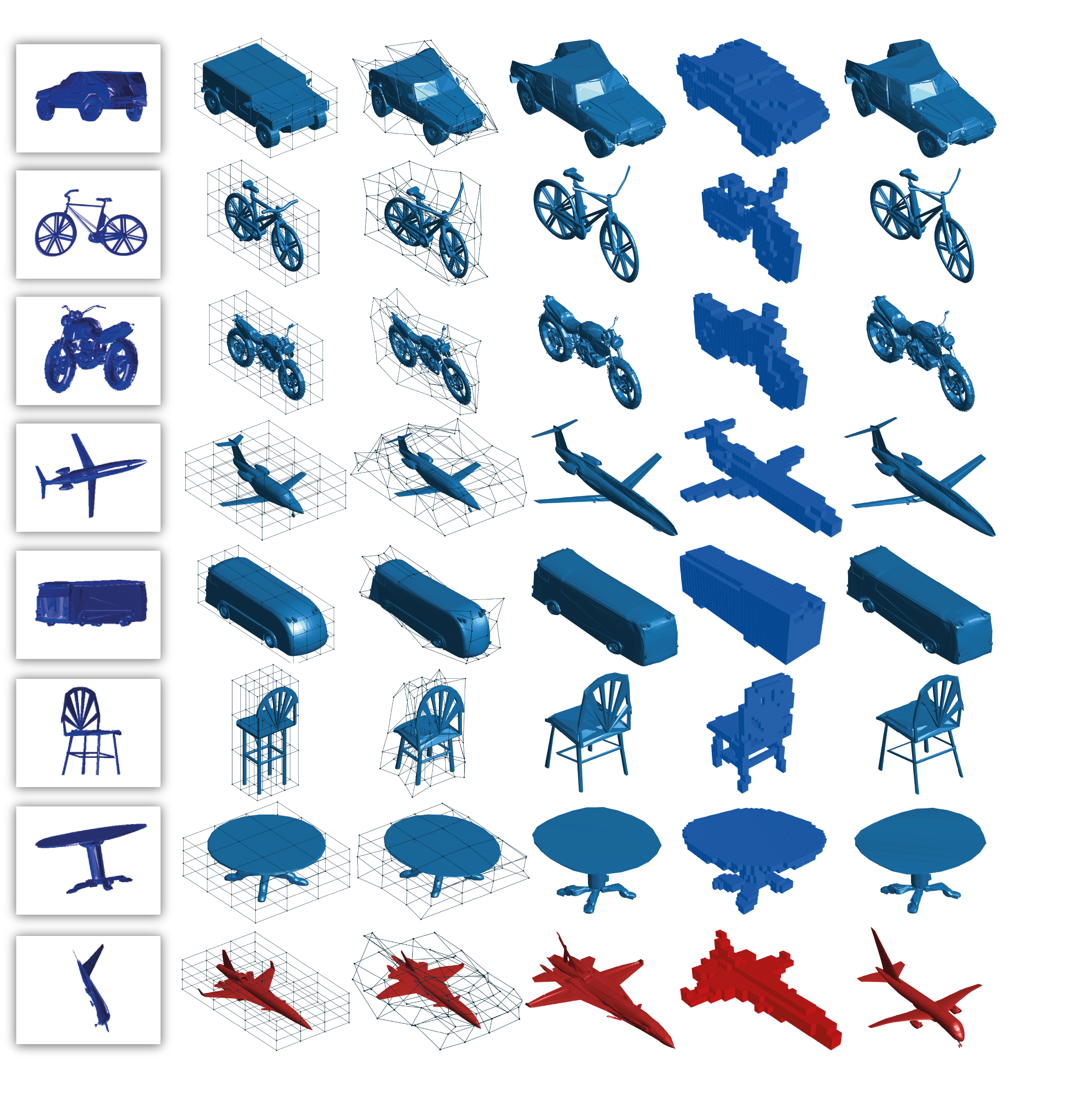}
\caption{Qualitative results for the synthetic dataset. Column (a) shows the input image. Column (b) shows the selected model from the graph. Column (c) shows the selected model with the \textsc{Ffd} parameters applied. The final 3D model reconstructed by applying the linear combination parameters is shown in column (d). The voxelized final model is shown in column (e) and the ground truth in column (f). In the success cases (blues), one can see that the final models are similar to the ground truth with slightly differences that can be hard to point it out. A failure case is shown on the last row in red where a ``wrong'' model was selected from the graph.}
\label{fig:bigPlot}
\end{figure*}

Qualitative results are shown in Figure~\ref{fig:bigPlot}. Column (a) shows the input synthetic image. Column (b) shows the selected model from the corresponding graph. Column (c) shows the selected model with the \textsc{Ffd} parameters applied. The final 3D model reconstructed by applying the linear combination parameters is shown in column (d). We also voxelize the final model for comparison reasons and it is shown in column (e). Ground truth is shown in column (f). Our proposed learning framework performed well at 
selecting a proper model to start the deformation process, and also at estimating the shape parameters to obtain the final deformed mesh model. In the successful cases, one can see that the final models are similar to the ground truth with slightly differences that can be hard to point it out. An interesting example to show the expressiveness of our proposed method is the chair case. One can note that the selected chair has long legs, but the estimated \textsc{Ffd} parameters managed to deform the chair to get shorter legs before applying the linear combination parameters to get the final model. A failure case is shown on the last row in red where an ``incorrect'' model was selected from the graph, in this case a fighter jet instead of a commercial airplane. 
This can in part be explained by the challenging image perspective of this instance. 
Even for a human it is difficult to correctly classify such an image, in this case a fighter jet is in 
fact a highly plausible choice of model\footnote{More results, failure cases, and videos can be found in the supplementary material or \href{url}{www.jhonykaesemodel.com}.}.

\begin{table}[ht!]
\centering\small
\resizebox{0.7\linewidth}{!}{\begin{tabular}{@{}rccccc@{}} 
\toprule
&\multicolumn{2}{c}{\textbf{\cite{Pontes2017}}} &\hphantom &\multicolumn{2}{c}{\textbf{Ours}}  \\
\cmidrule{2-3} \cmidrule{5-6} 
&$dist_{3D}$ &$IoU$  &&$dist_{3D}$ &$IoU$ \\ \midrule

\textbf{car} 		&\textbf{0.174} &\textbf{0.382}   &&0.179 &0.371  \\
\textbf{bicycle} 	&0.290 &\textbf{0.419}   &&\textbf{0.282} &0.402  \\
\textbf{motorbike}	&\textbf{0.084} &\textbf{0.384}   &&0.186 &0.309  \\
\textbf{aeroplane}	&0.262 &\textbf{0.442}   &&\textbf{0.153} &0.366  \\
\textbf{bus}		&0.091 &\textbf{0.376}   &&\textbf{0.058} &0.280  \\
\textbf{chair}		&\textbf{0.309} &\textbf{0.261}   &&0.461 &0.236  \\
\textbf{diningtable}&\textbf{0.353} &\textbf{0.256}   &&0.695 &0.223  \\
\textbf{sofa}		&\textbf{0.346} &\textbf{0.241}  &&0.573 &0.207  \\ \midrule
\textbf{Mean}		&\textbf{0.239} &\textbf{0.345}	&&0.323 &0.299  \\	

\bottomrule
\end{tabular}}
\caption{Quantitative results for our learning framework and for the method presented in \cite{Pontes2017} evaluated on the PASCAL3D+ dataset.}
\label{tab:results_pascal}
\end{table}

\textbf{3D Reconstruction from Real World Images.} The performance of our framework on the synthetic dataset was impressive. So we wanted to verify its generalization on a dataset with real world images. We evaluated the performance of the proposed method on the PASCAL3D+ dataset and we compared to the results presented in \cite{Pontes2017}. We found that it is a fair comparison since we are not playing with volumetric or point cloud representations but with dense polygonal meshes. In order to forward pass the real world images to our learning framework, we removed the image background since our framework was trained on images with white background. Table~\ref{tab:results_pascal} summarizes the results of our method and the results presented in \cite{Pontes2017} in terms of the surface distance, $dist_{3D}$, and the IoU. Our method did not outperform the method proposed in \cite{Pontes2017}, except for some classes. However, we achieved a similar performance on the real world dataset without using landmarks, silhouettes and the geometrical optimizations proposed in \cite{Pontes2017}. Moreover, since the ground truth models in the PASCAL3D+ dataset were chosen by humans to align it to images, the comparison metrics are not robust. As stated in \cite{Pontes2017}, most of the 3D reconstructions look closer to the images than the ground truth models themselves. This explains the high values for the surface distance and the low values for the IoU.

\begin{figure*}[ht!]
\centering
\textbf{\hspace{-23pt} (a) Input \hspace{8pt} (b) Selected model \hspace{8pt} (c) \textsc{Ffd} \hspace{13pt} (d) Final model \hspace{20pt} (e) \cite{Pontes2017} \hspace{23pt} (f) GT}
\includegraphics[trim={0cm 0.5cm 1cm 0.5cm},clip,width=0.83\textwidth,keepaspectratio]{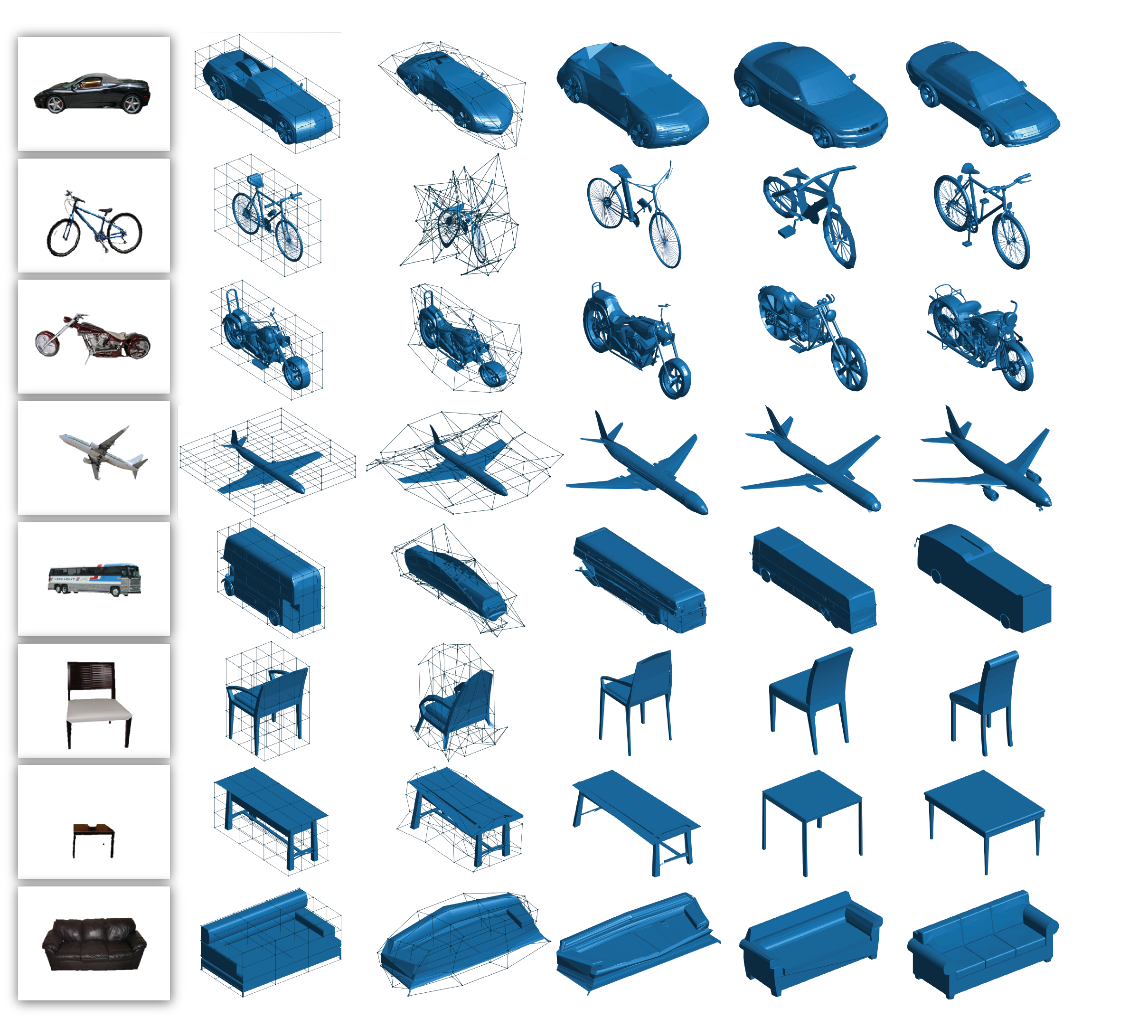}
\caption{Qualitative results for the real-world dataset. Column (a) shows the input image. Column (b) shows the selected model. Column (c) shows the selected model deformed by the \textsc{Ffd} parameters. The final 3D model reconstructed by applying the linear combination parameters is shown in column (d). We compare with \cite{Pontes2017} in column (e) and the ground truth is shown in column (f).}
\label{fig:results_pascal}
\end{figure*}

Qualitative results are shown in Figure~\ref{fig:results_pascal}. Column (a) shows the input image. Column (b) shows the selected model. Column (c) shows the selected model deformed by the \textsc{Ffd} parameters. The final 3D model reconstructed by applying the linear combination parameters is shown in column (d). We compare with \cite{Pontes2017} in column (e) and the ground truth is shown in column (f). One can see that our proposed method performed well on a real-world dataset. In the motorbike example it is clear when looking at the image that the motorbike does not have a backrest. The selected model was a good choice, since it shares topological similarities and although it has a backrest device, the linear combination step managed to diminish it. Another interesting example is the airplane where the selected model has a different type of wings, but the deforming process made it looks closer to the image.

\subsection{Implementation Details \& Limitations}
We designed our networks using PyTorch on a GPU Nvidia Tesla M40. We trained the convolutional autoencoder using a MSE loss function and the Adam optimizer with a learning rate of $1e^{-3}$ and a weight decay of $1e^{-5}$ during 100 epochs. To train the multi-label classifier we used a multi-label soft margin loss function and for the feedforward neural network we trained it using a MSE loss function. For both models we used the Adam optimizer with a learning rate of $1e^{-3}$ during 1,000 epochs.

One of the main limitations, which our method inherits from \cite{Pontes2017}, is the need for a good embedding graph. One can see in Figure~\ref{fig:fig2} that some models for the linear combination step have some crinkles on its surfaces. This happens during the graph creation when searching for dense correspondences between the models. This is especially important for the 3D reconstruction to get high quality models. However, finding dense correspondences between two different models that do not share the same number of vertices is still an open problem \cite{Pontes2017}. Another limitation is regarding the synthetic images since we do not synthesise a background that limits our work to white backgrounds. GANs can fit in this context to generate more realistic images.


\section{Conclusion}
We have proposed a simple yet effective learning framework to infer 3D mesh models from a single image using a compact mesh representation. A 3D mesh object is embedded in a low-dimensional space that allows one to perform model deformations by \textsc{Ffd} and sparse linear combination using a graph embedding structure. We demonstrated through experiments on synthetic and real-world datasets that our method convincingly reconstructs 3D mesh models from only a single image (\ie not using silhouettes and semantic landmarks) with fine-scaled geometry not yet achieved in previous works that rely on volumetric and point cloud representations. We believe that this work is a great first step towards more effective mesh representations for 3D geometric learning purposes.

\clearpage
{\small
\bibliographystyle{ieee}
\bibliography{egbib}
}

\end{document}